\documentclass[graybox]{svmult}

\usepackage{mathptmx}       % selects Times Roman as basic font
\usepackage{helvet}         % selects Helvetica as sans-serif font
\usepackage{courier}        % selects Courier as typewriter font
\usepackage{type1cm}        % activate if the above 3 fonts are
\usepackage{makeidx}         % allows index generation
\usepackage{graphicx}        % standard LaTeX graphics tool
\usepackage{multicol}        % used for the two-column index
\usepackage[bottom]{footmisc}% places footnotes at page bottom

\makeindex             % used for the subject index
\usepackage[T1]{fontenc}
\usepackage[utf8]{inputenc}

\usepackage{amsmath}
\usepackage{amssymb}

\newcommand{\MyTitle}{Map-aided Fusion Using Evidential Grids for Mobile Perception in Urban Environment}
\newcommand{\MyKeywords}{autonomous vehicle, autonomous driving, prior knowledge, dynamic fusion, contextual fusion, contextual discounting}
\newcommand{\MyAuthors}{Marek Kurdej, Julien Moras, V\'{e}ronique Cherfaoui, Philippe Bonnifait}

\usepackage{hyperref}
\hypersetup{
	bookmarksnumbered=true,
	bookmarksopen=true,
	bookmarksopenlevel=2,
	colorlinks=true, % false=use boxes, true=colors
	linkcolor=black,
	citecolor=black, %magenta,
	filecolor=black,
	urlcolor=blue,
	pdfinfo={  
	    Title={\MyTitle},
        Keywords={\MyKeywords},
	    Author={\MyAuthors},
	  } 
}

\usepackage{url}
\usepackage{subfigure}

\newcommand{\oppause}{\:}
\newcommand{\opconj}{\oppause\textcircled{\scriptsize{$\cap$}}\oppause}
\newcommand{\opdempster}{\oppause\oplus\oppause}
\newcommand{\opdisj}{\oppause\textcircled{\scriptsize{$\cup$}}\oppause}

\begin{document}

\title*{\MyTitle}
%\titlerunning{\MyTitle}

\author{\MyAuthors}

\institute{Marek Kurdej \email{marek.kurdej@hds.utc.fr}\and V\'{e}ronique Cherfaoui
\and Julien Moras \and Philippe Bonnifait \at UMR CNRS 6599 Heudiasyc,
University of Technology of Compi\`{e}gne, France}

\maketitle

\abstract*{
Evidential grids have been recently used for mobile object perception.
The novelty of this article is to propose a perception scheme using prior map knowledge.
A geographic map is considered an additional source of information fused with a grid representing sensor data.
Yager's rule is adapted to exploit the Dempster-Shafer conflict information at large.
In order to distinguish stationary and mobile objects, a counter is introduced and used as a factor for mass function specialisation.
Contextual discounting is used, since we assume that different pieces of information become obsolete at different rates.
Tests on real-world data are also presented.
}

\abstract{
Evidential grids have been recently used for mobile object perception.
The novelty of this article is to propose a perception scheme using prior map knowledge.
A geographic map is considered an additional source of information fused with a grid representing sensor data.
Yager's rule is adapted to exploit the Dempster-Shafer conflict information at large.
In order to distinguish stationary and mobile objects, a counter is introduced and used as a factor for mass function specialisation.
Contextual discounting is used, since we assume that different pieces of information become obsolete at different rates.
Tests on real-world data are also presented.
}

\section{Introduction}
\label{sec:introduction}

Autonomous driving has been an important challenge in recent years.
Navigation and precise localisation aside, environment perception is an important on-board system of a self-driven vehicle.
The level of difficulty in autonomous driving increases in urban environments, where a good scene understanding makes the perception subsystem crucial.
There are several reasons that make cities a demanding environment.
Poor satellite visibility deteriorates the precision of GPS positioning.
Vehicle trajectories are hard to predict due to high variation in speed and direction.
Also, the sheer number of mobile objects poses a problem, e.g. for tracking algorithms.

On the other hand, more and more detailed and precise geographic databases become available.
This source of information has not been well examined yet, hence our approach of incorporating prior knowledge from digital maps in order to improve perception scheme.
A substantial amount of research has focused on the mapping problem for autonomous vehicles,
e.g. Simultaneous Localisation and Mapping (SLAM) approach, but the use of maps for perception is still understudied.

In this article, we propose a data fusion method based on Dempster--Shafer theory \cite{Shafer1976} taking into account meta-knowledge obtained from a digital map.
We show the advantage of including prior knowledge into an embedded perception system of an autonomous car. The vehicle environment is modelled by 2D occupancy grids proposed in \cite{Elfes1989}.
This paper describes a robust and unified approach to a variety of problems in spatial representation using the theory of probability.
The theory of evidence was not combined with occupancy grids until recently to build environment maps for robot perception \cite{Pagac1998}.
Only recent works take advantage of the theory of evidence in the context of mobile perception \cite{Moras2010b}.
Some works use 3D city model as a source of prior knowledge for localisation and vision-based perception \cite{Cappelle2011}, whereas our method uses maps for scene understanding.

This article is organised as follows. In section \ref{sec:method}, we describe the details of the method. % starting with the description of data sources and the purpose of each grid. Further, details on the information fusion are given.
Section \ref{sec:results} presents the results and section \ref{sec:conclusion} concludes the paper.

%\vspace{-.5cm}

\section{Multi-grid fusion approach}
\label{sec:method}

\begin{figure}
	\centering
	\includegraphics[width=1\textwidth]{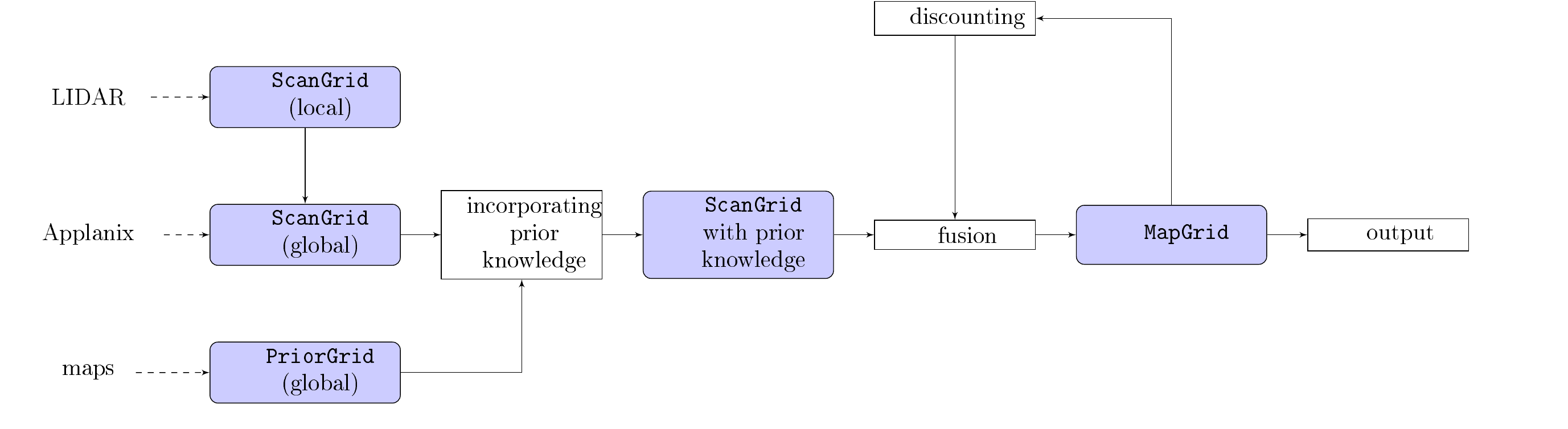}
	\caption{Method overview (lidar: laser scanner, Applanix: inertial measurement unit).}
	\label{fig:fusion-general}
\end{figure}

This section presents the proposed perception schemes.
The grid construction method is described in section \ref{sub:grids} and all data processing steps are detailed in section \ref{sub:fusion}.
Figure \ref{fig:fusion-general} presents a general overview of our approach.

%\vspace{-.5cm}

%\vspace{-1cm}

\subsection{Heterogeneous data sources}
\label{sub:sources}

There are three sources in our perception system: vehicle pose, lidar range scanner point cloud and vector maps.
The vehicle pose comes from the Applanix system based on a GPS, an odometer and an IMU.
The system is supposed to provide precise and integral positioning.
Our main source of information about the environment is an IBEO Alaska XT lidar able to provide a cloud of about 800 points 10 times per second.
The digital maps that we use were provided by the French National Geographic Institute (IGN) and contain 3D building models as well as the road surface.
We also performed successful tests with freely available OpenStreetMap project 2D maps \cite{OpenStreetMap}, but here we limited the use to building data.
We assume the maps to be precise and accurate.

\subsection{Occupancy grids}
\label{sub:grids}

An occupancy grid models the world using a tessellated representation of spatial information.
In general, it is a multidimensional spatial lattice with cells storing some stochastic information.
In our case, each cell representing a box (a part of environment) $X\times Y$ where $X=\left[x_{-},\, x_{+}\right]$, $Y=\left[y_{-},\, y_{+}\right]$ stores a mass function.

\begin{itemize}
\item \texttt{ScanGrid} (SG) construction: In order to process the lidar data, an evidential occupancy grid is computed when a new scan arrives, this grid is called \texttt{ScanGrid}. Each cell of this grid stores a mass function on the frame of discernment (FOD) $\Omega_{SG}=\left\{ F,O\right\} $, where $F$ refers to the free space and $O$ -- to the occupied space. The basic belief assignment, which reflects the sensor model, is described in \cite{Moras2010b}.
\item \texttt{MapGrid} (MG): To store the results of information fusion, an occupancy grid MG has been introduced with a FOD $\Omega_{MG}=\left\{ F,\, C,\, N,\, S,\, V\right\} $. Respective classes represent: free space $F$, mapped infrastructure (buildings) $C$, non-mapped infrastructure $N$, temporarily stopped objects $S$ and mobile (moving) $V$ objects. $\Omega_{MG}$ is a common frame used for information fusion. By using MG as a cumulative information storage, we are not obliged to aggregate preceding \texttt{ScanGrid}s.
\item \texttt{PriorGrid} (PG) context representation:
PG allows us to perform a contextual information fusion incorporating some meta-knowledge about the environment.
This grid uses the same frame of discernment $\Omega_{MG}$ as MG.
The grid is obtained by projection of map data, buildings and roads, onto a 2D grid with global coordinates.
\\
We define two sets of polygons defining the 2D position of buildings and road surface by, respectively, $\mathcal{B}=\left\lbrace b_{i}=\begin{bmatrix}x_{1}x_{2}\ldots x_{m_{i}}\\
y_{1}y_{2}\ldots y_{m_{i}}
\end{bmatrix},i\in[0,n_{B}]\right\rbrace $ and $\mathcal{R}=\left\lbrace r_{i}=\begin{bmatrix}x_{1}x_{2}\ldots x_{m_{i}}\\
y_{1}y_{2}\ldots y_{m_{i}}
\end{bmatrix},i\in[0,n_{R}]\right\rbrace $, $\mathcal{B}\cap\mathcal{R}=\emptyset$.
Then, we attribute the mass to each cell $\left\{ X,Y\right\}$ of the \texttt{PriorGrid} in the
following way:\\
We note that $B=\left\{ C\right\} $, $R=\left\{ F,\, S,\, V\right\} $, $T=\left\{ F,\, N,\, S,\, V\right\} $ for convenience and readability only.
$A$ denotes all other strict subsets of $\Omega$.
These aliases characterise the meta-information inferred from geographic maps.
For instance, on the road surface $R$, we $encourage$ the existence of free space $F$ as well as stopped $S$ and moving $V$ objects.
Analogically, building information $B$ fosters mass transfer to $C$.
Lastly, $T$ denotes the intermediate area, e.g. pavements, where mobile and stationary objects as well as small urban infrastructure can be present.
Note that neither buildings nor roads are present, so we exclude existence of mapped infrastructure $C$, but we cannot omit other classes.
Also, we define a level of confidence $\beta$ for each map source, possibly different for each context.
Let $\tilde{x}=\frac{x_{-}+x_{+}}{2}$, $\tilde{y}=\frac{y_{-}+y_{+}}{2}$.
\end{itemize}
\begin{align}
	m_{PG}\{X,Y\}(B) & =\begin{cases}
	\beta_{B} & \text{if }(\tilde{x},\tilde{y})\in b_{i}\\
	0 & \text{otherwise}
	\end{cases} & \forall i\in[0,n_{B}]
	\nonumber
	\\
	m_{PG}\{X,Y\}(R) & =\begin{cases}
	\beta_{R} & \text{if }(\tilde{x},\tilde{y})\in r_{i}\\
	0 & \text{otherwise}
	\end{cases} & \forall i\in[0,n_{R}]
	\nonumber
	\\
	m_{PG}\{X,Y\}(T) & =\begin{cases}
	0 & \text{if }(\tilde{x},\tilde{y})\in b_{i}\vee(\tilde{x},\tilde{y})\in r_{j}\\
	\beta_{T} & \text{otherwise}
	\end{cases} & \forall i\in[0,n_{B}],\forall j\in[0,n_{R}]
	\nonumber
	\\
	m_{PG}\{X,Y\}(\Omega) & =\begin{cases}
	1-\beta_{B} & \text{if }(\tilde{x},\tilde{y})\in b_{i}\\
	1-\beta_{R} & \text{if }(\tilde{x},\tilde{y})\in r_{i}\\
	1-\beta_{T} & \text{otherwise}
	\end{cases} & \forall i\in[0,n_{B}],\forall j\in[0,n_{R}]
	\nonumber
	\\
	m_{PG}\{X,Y\}(A) & =0 & \forall A\subsetneq\Omega \text{ and } A\notin\left\{ B,R,T\right\} 
\end{align}
%Common sense provides us with useful information about what type of objects can be found in various contexts.

\subsection{Incorporating prior knowledge}
\label{sub:prior}

The frame of discernment $\Omega_{SG}$ used in SG is distinct from
$\Omega_{MG}$, so in order to enable the fusion of SG and MG we define
a refining $r_{SG}\,:\,2^{\Omega_{SG}}\rightarrow2^{\Omega_{MG}}$
such that $r_{SG}\left(\left\{ F\right\} \right)=\left\{ F\right\} $,
$r_{SG}\left(\left\{ O\right\} \right)=\left\{ C,N,S,V\right\} $,
$r_{SG}(A)=\bigcup_{\theta\in A}r_{SG}(\theta)$. The refined mass
function can be expressed as $m_{SG}^{\Omega_{MG}}\left(r_{SG}\left(A\right)\right)=m_{SG}^{\Omega_{SG}}\left(A\right),\,\forall A\subseteq\Omega_{SG}$.
Then, Dempster's rule is applied in order to exploit the prior information
included in \texttt{PriorGrid}: 
\begin{align}
	m'{}_{SG,\, t}^{\Omega_{MG}} & =m_{SG,\, t}^{\Omega_{MG}}\opdempster m_{PG}^{\Omega_{MG}}
\end{align}

\subsection{Temporal fusion}
\label{sub:fusion}

\subsubsection*{Computing conflict masses}
\label{sub:conflict}

We use the idea from \cite{Moras2011} to distinguish between two types of conflict,
which arise from the fact that the environment is dynamic.
We denote $\emptyset_{FO}$ the conflict induced when a free cell in MG is fused with an occupied cell in SG.
Similarly, $\emptyset_{OF}$ indicates the conflicted caused by an occupied cell
in MG fused with a free cell in SG. In an error-free case, these conflicts
represent, respectively, the disappearance and the appearance of an object.
Conflict masses are calculated using the formulas: $m_{MG,\, t}\left(\emptyset_{OF}\right)=m_{MG,\, t-1}\left(O\right)\cdot m_{SG,\, t}\left(F\right)$,
$m_{MG,\, t}\left(\emptyset_{FO}\right)=m_{MG,\, t-1}\left(F\right)\cdot m_{SG,\, t}\left(O\right)$,
where $m(O)=\sum\limits _{A}m(A),\;\forall A\subseteq\left\{ C,N,S,V\right\} $.

\subsubsection*{\texttt{MapGrid} specialisation using a counter}
\label{sub:specialisation}

Mobile object detection is an important issue in dynamic environments.
We propose the introduction of a counter $\zeta$ in each cell in order to include temporal information on the cell occupancy.
For this purpose, incrementation and decrementation steps $\delta_{inc}\in[0,1]$,
$\delta_{dec}\in[0,1]$, as well as threshold values $\gamma_{O}$,
$\gamma_{\emptyset}$ have been defined.
\begin{align}
	\zeta^{(t)} & =\min\left(1,\,\zeta^{(t-1)}+\delta_{inc}\right) & \text{if }m_{MG}(O)\geq\gamma_{O} \text{ and } m_{MG}\left(\emptyset_{FO}\right)+m_{MG}\left(\emptyset_{OF}\right)\leq\gamma_{\emptyset}
	\nonumber
	\\
	\zeta^{(t)} & = \max\left(0,\,\zeta^{(t-1)}-\delta_{dec}\right) & \text{if } m_{MG} \left( \emptyset_{FO} \right) + m_{MG} \left(\emptyset_{OF}\right)>\gamma_{\emptyset}
	\nonumber
%	\\
%	\zeta^{(t)} & = \zeta^{(t-1)} & \text{ otherwise}
\end{align}

Otherwise $\zeta(t)$ rests unchanged.
Using $\zeta$ values, we impose a specialisation of mass functions
in MG using the equation:
\begin{align}
	m'{}_{MG,\, t} & \left(A\right)=S(A,B)\cdot m_{MG,\, t}(B)
\end{align}
where specialisation matrix $S(\cdot,\cdot)$ is defined as:

\begin{align}
	S(A\backslash\left\{ V\right\} ,\, A) & =\zeta & \forall A\subseteq\Omega_{MG} \text{ and } \left\{ V\right\} \in A\nonumber \\
	S(A,\, A) & =1-\zeta & \forall A\subseteq\Omega_{MG} \text{ and } \left\{ V\right\} \in A\nonumber \\
	S(A,\, A) & =1 & \forall A\subseteq\Omega_{MG} \text{ and } \left\{ V\right\} \notin A\nonumber \\
	S(\cdot,\,\cdot) & =0 & \text{otherwise}
\end{align}

\subsubsection*{Fusion rule}
\label{sub:fusion-rule}

An important part of the method consists in fusing a discounted and specialized
MG (see section \ref{sub:discounting} and preceding paragraph) with a SG combined with prior
knowledge (see section \ref{sub:prior}).
\begin{equation}
	m_{MG,\, t}=^{\alpha}m'{}_{MG,\, t-1}\circledast m'{}_{SG,\, t}
	\label{eq:fusion-general}
\end{equation}

The fusion rule $\circledast$ is a modified Yager's rule \cite{Yager1987} adapted to mobile object detection.
There are of course many different rules that could be used, but in order to distinguish between moving and stationary objects some modifications had to be included.
These modifications consist in transferring the mass corresponding to a newly appeared object $\emptyset_{FO}$ to the class of moving objects $V$ as described by the equation \ref{eq:fusion-yager-modified}.
Symbol $\opconj$ denotes the conjunctive fusion rule.
\begin{align}
	\left(m_{1}\circledast m_{2}\right)\left(A\right) & =\left(m_{1}\opconj m_{2}\right)(A) & \forall A\subsetneq\Omega\wedge A\neq V\nonumber \\
	\left(m_{1}\circledast m_{2}\right)\left(V\right) & =\left(m_{1}\opconj m_{2}\right)\left(V\right)+\left(m_{1}\opconj m_{2}\right)\left(\emptyset_{FO}\right)\nonumber \\
	\left(m_{1}\circledast m_{2}\right)\left(\Omega\right) & =\left(m_{1}\opconj m_{2}\right)\left(\Omega\right)+\left(m_{1}\opconj m_{2}\right)\left(\emptyset_{OF}\right)\nonumber \\
	\left(m_{1}\circledast m_{2}\right)\left(\emptyset_{FO}\right) & =0\nonumber \\
	\left(m_{1}\circledast m_{2}\right)\left(\emptyset_{OF}\right) & =0
	\label{eq:fusion-yager-modified}
\end{align}

All the above steps allow us to construct a \texttt{MapGrid} containing reach information on the environment state, including the knowledge on mobile and static objects.

\subsection{Contextual discounting}
\label{sub:discounting}

%TODO: allow STH to do sth
Information discounting allows to \emph{forget} information which is no longer valid.
Discounting parameter $\mathbf{\alpha}$ serves to model the speed with which information becomes obsolete.
Thanks to the contextual discounting \cite{Mercier2008}, we make use of more
detailed information regarding the confidence we have in the source in various contexts.
We noticed that different pieces of information become obsolete with different speed.
Hence, the coarsening used is $\Theta=\left\{ \theta_{static},\,\theta_{dynamic},\,\theta_{free}\right\} $,
with $\theta_{static}=\left\{ C,N\right\} $, $\theta_{dynamic}=\left\{ S,V\right\} $,
$\theta_{free}=\left\{ F\right\} $, and discount rates $\mathbf{\alpha}=\left\{ \alpha_{static},\,\alpha_{dynamic},\,\alpha_{free}\right\}$.
We assign higher discount rates (lower confidence) to rapidly changing contexts such as free space, stopped and moving objects, and lower rates to the static context.
The discounted mass function is obtained by the disjunctive combination of the input mass function
$m_{MG}$ and mass functions for each element of the partition $\Theta$.

\begin{align}
	{}^{\alpha}m_{MG,\, t} & = m_{MG,\, t}\opdisj m_{static}\opdisj m_{dynamic}\opdisj m_{free}
\end{align}
where each mass function $m_{l}$ $(l=\text{\text{static},\text{ dynamic},\text{ \text{free}}})$
is defined by $m_{l}\left(\theta_{l}\right)=\alpha_{l}$, $m_{l}\left(\emptyset\right)=1-\alpha_{l}$,
$m_{l}(A)=0,\,\forall A\subseteq\Omega\wedge A\notin\left\{ \emptyset,\theta_{l}\right\} $.

\begin{figure}
	\centering
	\includegraphics[width=0.9\textwidth]{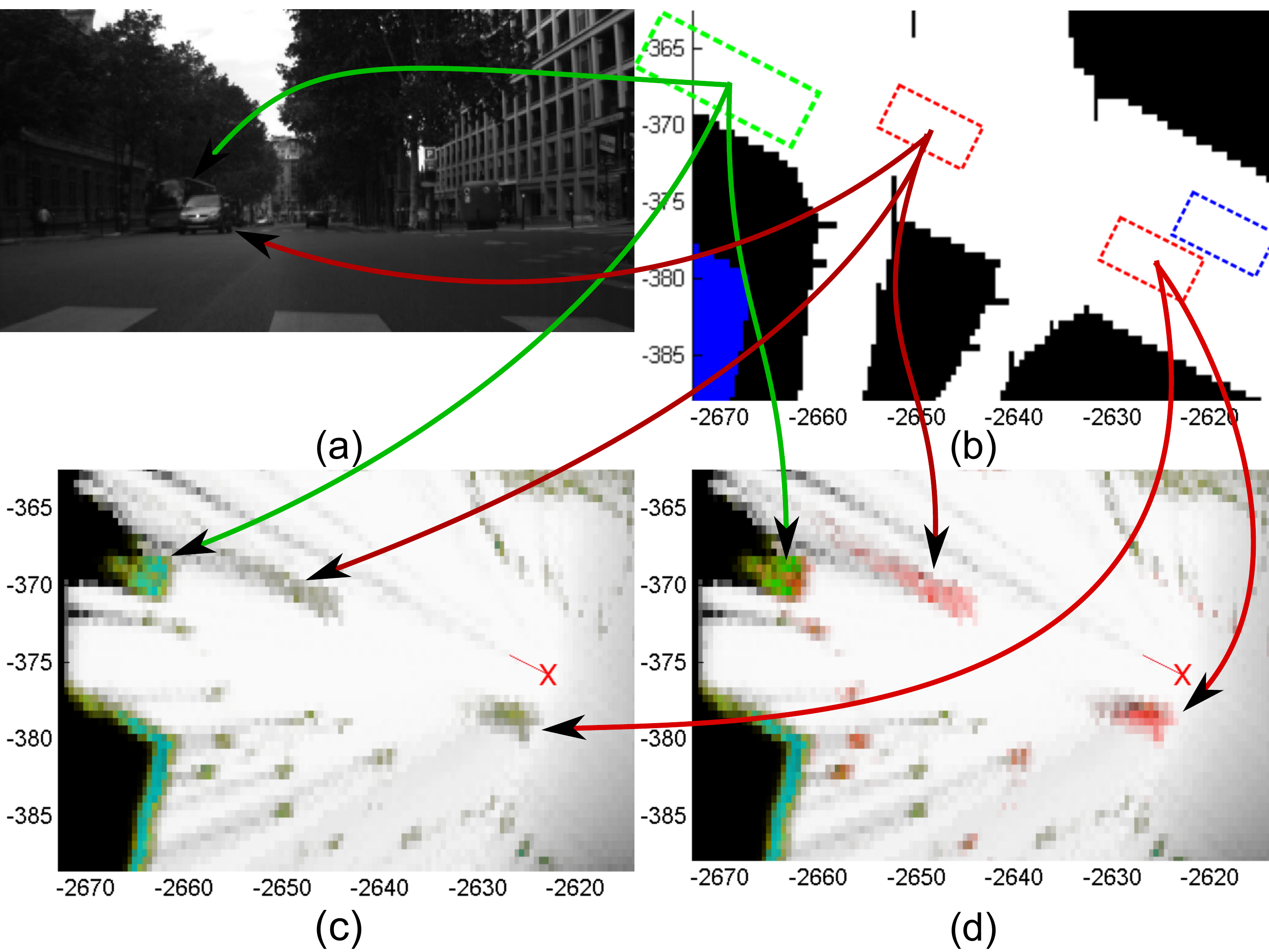}
%	\subfigure[] {
%		\includegraphics[width=0.48\textwidth]{figures/camera-bus1}
%		\label{fig:results-scene}
%	}
%	\subfigure[] {
%		\includegraphics[width=0.48\textwidth]{figures/prior1-annotated}
%		\label{fig:results-prior}
%	}
%	\\[-0.3cm]
%	\subfigure[] {
%%		\includegraphics[width=0.48\textwidth]{figures/classes-bus1-annotated}
%		\includegraphics[width=0.48\textwidth]{figures/classes-bus1}
%		\label{fig:results-grid-noprior}
%	}
%	\subfigure[] {
%%		\includegraphics[width=0.48\textwidth]{figures/complete-bus1-annotated}
%		\includegraphics[width=0.48\textwidth]{figures/complete-bus1}
%		\label{fig:results-grid}
%	}
%	\vspace{-0.3cm}
%	\caption{\subref{fig:results-scene} Scene. \subref{fig:results-prior} PG. \subref{fig:results-grid-noprior} MG without prior information. \subref{fig:results-grid} MG with prior map knowledge.}
	\caption{(a) Scene. (b) PG. (c) MG without prior information. (d) MG with prior map knowledge.}
	\label{fig:results}
\end{figure}

\section{Results}
\label{sec:results}

\subsection{Setup}
\label{sub:setup}

The data set used for our experiments was acquired in cooperation with IGN in Paris.
The overall length of the trajectory was about 3~km.
The size of the grid cell in the occupancy grids was set to 0.5~m, which is sufficient to model a complex environment with mobile objects.
The discount rates $\mathbf{\alpha}$ describing the speed of information becoming obsolete were defined empirically, but they can be learnt from data, as proposed in \cite{Mercier2008}.
We have defined the map confidence factor $\mathbf{\beta}$ by ourselves, but ideally, it should be given by the map provider.
$\beta$ describes data currentness (age), errors introduced by geometry simplification and spatial discretisation.
$\mathbf{\beta}$ can also be used to depict the localisation accuracy.
Other parameters, such as counter steps $\delta_{inc}$, $\delta_{dec}$ and thresholds $\gamma_{O}$, $\gamma_{\emptyset}$ used for mobile object detection determine the sensitiveness of mobile object detection and were set by manual tuning.

\subsection{Impact of prior knowledge}
\label{sub:results-prior}

The results for a particular instant of the approach tested on real-world data are presented on figure \ref{fig:results}.
The visualisation of the MG has been obtained by calculating the pignistic probability of each class \cite{Smets2005}.
The presented scene contains two cars (only one is visible in the camera image) going in the direction opposite to the test
vehicle and a bus parked on the road edge.
Bus and car positions are marked on the grids by green and red boxes, respectively.
The test vehicle position is shown as a blue box.
Different classes of $\Omega_{MG}$ are represented by different colours: $F$ -- white, $C,\, N$ -- blue, $S$ -- green and $V$ -- red.
PG on figure \ref{fig:results}(b) shows the position of the road space (white) and buildings (blue).

The principal advantage gained by using map knowledge is richer information on the detected objects.
A clear difference between a moving object (red, car) and a stopped one (green, bus) is visible.
Also, stopped objects are distinct from infrastructure when prior map information is available (cf. figures \ref{fig:results}(c) and \ref{fig:results}(d).
In addition, thanks to the prior knowledge, stationary objects (cyan) such as infrastructure are distinguished from stopped objects on the road.
Grids make noticeable the effect of discounting, as information on the environment behind the vehicle is being forgotten.
On the other hand, the parked bus is still in evidence despite being occluded by the passing car.

\section{Conclusion and perspectives}
\label{sec:conclusion}

A new mobile perception scheme based on prior map knowledge has been introduced.
Geographic information is exploited to reduce the number of possible hypotheses delivered by an exteroceptive source.
A modified fusion rule taking into account the existence of mobile objects has been defined.
Furthermore, the variation in information lifetime has been modelled by the introduction of contextual discounting.
In the future, we anticipate removing the hypothesis that the map is accurate.
This approach will entail considerable work on creating appropriate error models for the data source.
Moreover, we envision differentiating the free space class into two complementary classes to distinguish navigable and non-navigable space.
This will be a step towards the use of our approach in autonomous navigation.
Another perspective is the use of reference data to validate the results, choose the most appropriate fusion rule and learn algorithm parameters.
We envision using map information to predict object movements.
It rests also a future work to exploit fully the 3D map information.

\begin{acknowledgement}
	This work has been supported by ANR (French National Agency) CityVIP project under grant {ANR-07\_TSFA-013-01}.
\end{acknowledgement}

\end{document}